\documentclass[journal]{IEEEtran}
 \usepackage{amsmath}
 \usepackage[caption=false,font=normalsize,labelfont=sf,textfont=sf]{subfig}
 \usepackage{graphicx}
\begin{document}

\title{Kolmogorov–Arnold Networks for Spatially\\ Independent Multispectral Land Classification}

\author{Katherine L. Bauer $^{1, *}$, Teemu Härkönen $^{2}$, Simo Särkkä $^{2}$, Arturo Sanchez-Azofeifa $^{1}$

\begin{center}
    $^{1}$Department of Earth and Atmospheric Sciences, University of Alberta, Edmonton, AB, Canada\\
    $^{2}$Department of Electrical Engineering and Automation, Aalto University, Helsinki, Finland\\
    $^{*}$Corresponding author: klbauer1@ualberta.ca\\
    
\end{center}

}



\maketitle

\begin{abstract}

Land classification from satellite imagery is important for land management, environmental monitoring, and urban planning. Machine learning methods such as random forests and multilayer perceptrons have shown strong performance on multispectral data, while the Kolmogorov-Arnold network has emerged as an alternative architecture with compact model structures. This study evaluates the Kolmogorov-Arnold network for land classification using Landsat 8 imagery and compares it with random forest and multilayer perceptron models. The models were trained and tested on data from Edmonton, Alberta and evaluated on an independent dataset from Calgary, Alberta across five land classes: agriculture, urban, water, forest, and bare ground. For the Calgary dataset, the Kolmogorov-Arnold network matched the accuracy of the random forest and outperformed the multilayer perceptron, while requiring substantially fewer trainable parameters and providing greater interpretability.

\end{abstract}


\section{Introduction}

\IEEEPARstart{C}{lassifying} land types using satellite data is important for informed decision-making processes in land management, resource allocation, environmental monitoring, and urban planning. This task involves assigning each pixel to a specific land use or land cover (LULC) category. \textit{Land use} describes how humans manage or manipulate the surface through means such as urban construction and agriculture, while \textit{land cover} refers to the physical material present on the Earth's surface, such as vegetation and water  \cite{martinez2012land}. Accurate LULC classification supports the analysis of land cover change, deforestation, agricultural activity, and natural resource distribution \cite{atef2023modelling, basheer2022comparison}. Traditional approaches can be labour-intensive and prone to subjectivity \cite{zhao2023land}. Multispectral satellite data, which captures reflectance in several broad wavelength bands, provides a lighter and more accessible alternative compared to hyperspectral data, which is often limited in availability. This is particularly useful when distinguishing expansive LULC classes across different areas. As remote sensing technologies continue to advance, integrating novel machine learning methods has become increasingly important for scalable, accurate, and automated land classification. 

Following the increasing adoption of machine learning in remote sensing, ensemble-based algorithms such as the
\IEEEpubidadjcol
random forest (RF), became a dominant approach for supervised classification \cite{zhang2022review}. RF builds multiple decision trees on bootstrap samples, aggregating their votes to boost accuracy and mitigate overfitting. To ensure stable predictions, RFs often require a relatively large number of parameters, with the suggested default number of trees being 500 \cite{belgiu2016random}. Since each tree contains multiple decision nodes and thresholds, this can result in models with high model complexity, depending on tree depth and structure. Moreover, the independence of each tree renders RF a “black-box” model, often necessitating post-hoc interpretability tools such as Shapley additive explanations (SHAP) to explain predictions \cite{kim2022shap}. Despite these drawbacks, RF remains an accepted choice for heterogeneous landscapes. 

As model architectures progressed, neural networks gained popularity for land classification. Among these, the multilayer perceptron (MLP), a class of feedforward neural network, offers a greater capacity for modeling intricate, non-linear patterns in spectral data. Like RFs, MLPs make no assumptions about the data distribution. Instead, they generate predictions by transforming the input through layers of interconnected nodes, each applying fixed activation functions and learned weights \cite{goodfellow2016deep}. MLPs can benefit from increased depth and width when modeling complex spectral relationships \cite{papoutsis2023benchmarking}. This in turn affects the interpretability of the MLP since deeper and more complex networks become increasingly difficult to analyze and understand, again requiring post-hoc interpretability methods similar to RFs. 

An emerging alternative to these models is the Kolmogorov-Arnold network (KAN), which shares structural similarities with MLPs but differs fundamentally in function representation \cite{liu2024kan}. Instead of fixed activation functions and learnable weights, KANs learn univariate spline functions on its edges, which are summed across nodes to approximate complex multivariate functions. This architecture is inspired by the Kolmogorov-Arnold representation theorem, which states that a complex, multivariate function can be broken down into the summation of several simple, univariate functions \cite{schmidt2021kolmogorov}. The original KAN paper demonstrated promising results primarily on small-scale scientific modeling tasks, and have the potential to achieve superior accuracy than the MLP while using fewer trainable parameters \cite{liu2024kan}. KANs also offer greater interpretability through their learnable univariate functions, which can be visualized and inspected directly without relying solely on post-hoc explanation tools \cite{bozorgasl2024wavelet}. The \textit{PyKAN} (v0.2.8) package supports this through built-in visualization and spline inspection capabilities \cite{liu2024kan}. 

These key attributes motivate the application of KANs to multispectral land classification, particularly for distinguishing similar LULC classes. Therefore, the objective of this study is to evaluate the performance of RF, MLP, and KAN architectures for multi-class LULC classification within two major Albertan urban centers, focusing on accuracy, efficiency, and interpretability. 

The contributions of this paper are as follows:
\begin{enumerate}
    \item The first cross-city KAN application for LULC classification;
    \item Evaluating differences in KAN performance when trained on a dataset consisting of normalized band values versus a dataset of selected spectral indices;
    \item Conducting a comparative analysis of parameter efficiency and interpretability of the KAN against the RF and MLP.
\end{enumerate}

These contributions directly address both methodological and applied gaps in the current LULC classification literature. By evaluating KANs in a cross-city urban setting, this study benchmarks and tests model generalization under realistic spatial conditions. Comparing a normalized band dataset with a spectral indices dataset clarifies how feature representation interacts with KAN's functional structure, providing practical guidance for model design in multispectral remote sensing. Finally, the systematic comparison of accuracy, parameter efficiency, and interpretability against established RF and MLP baselines positions KANs not only as a high-performing classifier, but as a viable, explainable alternative for urban land cover mapping where architectural compactness and transparency are critical for operational and policy-driven applications.

\section{Related Works}

The first work to apply the KAN to multispectral land classification combined it with a convolutional neural network and evaluated it on the EuroSAT dataset. The author tested the KAN's accuracy with 256 hidden nodes versus 32 hidden nodes and concluded that the 96\% accuracy remained consistent when the model's size was reduced, showing that smaller KAN models still achieve high accuracies \cite{cheon2024kolmogorov}.

Fawzy et al. tested different combinations of KAN hyperparameters for land classification in Budapest, Hungary, performing a grid search over the number of hidden layers, hidden nodes, grid size, and spline order ($k$). The authors found that a shallow model with a lower value of $k$ performed best, achieving an accuracy of 88.89\% across four LULC classes (buildings, roads, vegetation, and water), compared to 87.84\% for a benchmark shallow neural network \cite{fawzy2025vhr}. 

In the context of hybrid architectures, Han et al. combined the KAN with a Vision Transformer (ViT) for land classification using the quad-band Hainan PolSAR dataset. Each pixel value was explicitly min-max normalized to maintain consistency across images. The ViT-KAN achieved a higher accuracy of 96.24\% using 120.5k parameters compared to the ViT-MLP baseline, which achieved an accuracy of 92.85\% using 135.2k parameters \cite{han2025vit}. For hyperspectral land classification, Jamali et al. proposed HybridKAN, which integrates multiple 2D and 3D KAN models into a unified architecture, outperforming CNN- and ViT-based benchmarks \cite{jamali2024learn}. 

Martellozzo et al. examined how LULC changes affect soil quality and agricultural production in the Edmonton-Calgary corridor using a longitudinal study. A supervised maximum likelihood classifier was used, which ranged in accuracies between 87\% and 94\% \cite{martellozzo2015urbanization}. Similarly, Stan \& Sanchez explored the Edmonton-Calgary corridor over a 22 year period to examine LULC change under different theoretical policies. The paper used a decision tree-based classification method for each year, which achieved accuracies between 78\% and 82\% \cite{stan2017edmonton}.

Collectively, these studies establish that KANs are a competitive and parameter-efficient alternative to conventional neural architectures for LULC classification across multispectral and hyperspectral domains. Together, these findings motivate the present research by highlighting both the demonstrated strengths of KANs in remote sensing classification and a gap in their application to regional, policy-relevant LULC analysis in Edmonton and Calgary, thereby positioning this work as a methodological and applied extension of the existing literature. 

\section{Methods}

\subsection{Study Area}

Edmonton, Alberta, Canada is located on the North Saskatchewan River and is the center of the Edmonton Metropolitan area. The terrain of the Edmonton area is flat to gently rolling and mostly consists of remnant grasslands, sand dunes, and peat-land habitats. These ecosystems are characteristic of low cover, dry, greenish-brown vegetation. Through Edmonton runs the North Saskatchewan River Valley, which is a preserved ravine system featuring mainly mixed wood forest and wetland habitats. In 2024, Edmonton’s population reached 1.19 million \cite{govab_pop}.

Calgary, located 300 km south from Edmonton, lies at the transition zone between the Rocky Mountains and the Canadian Prairies, with the Bow and Elbow Rivers running through. As of 2024, Calgary’s population was 1.6 million \cite{calgarydashboard}. 

\begin{figure}[ht]
\centering
\includegraphics[width=\linewidth]{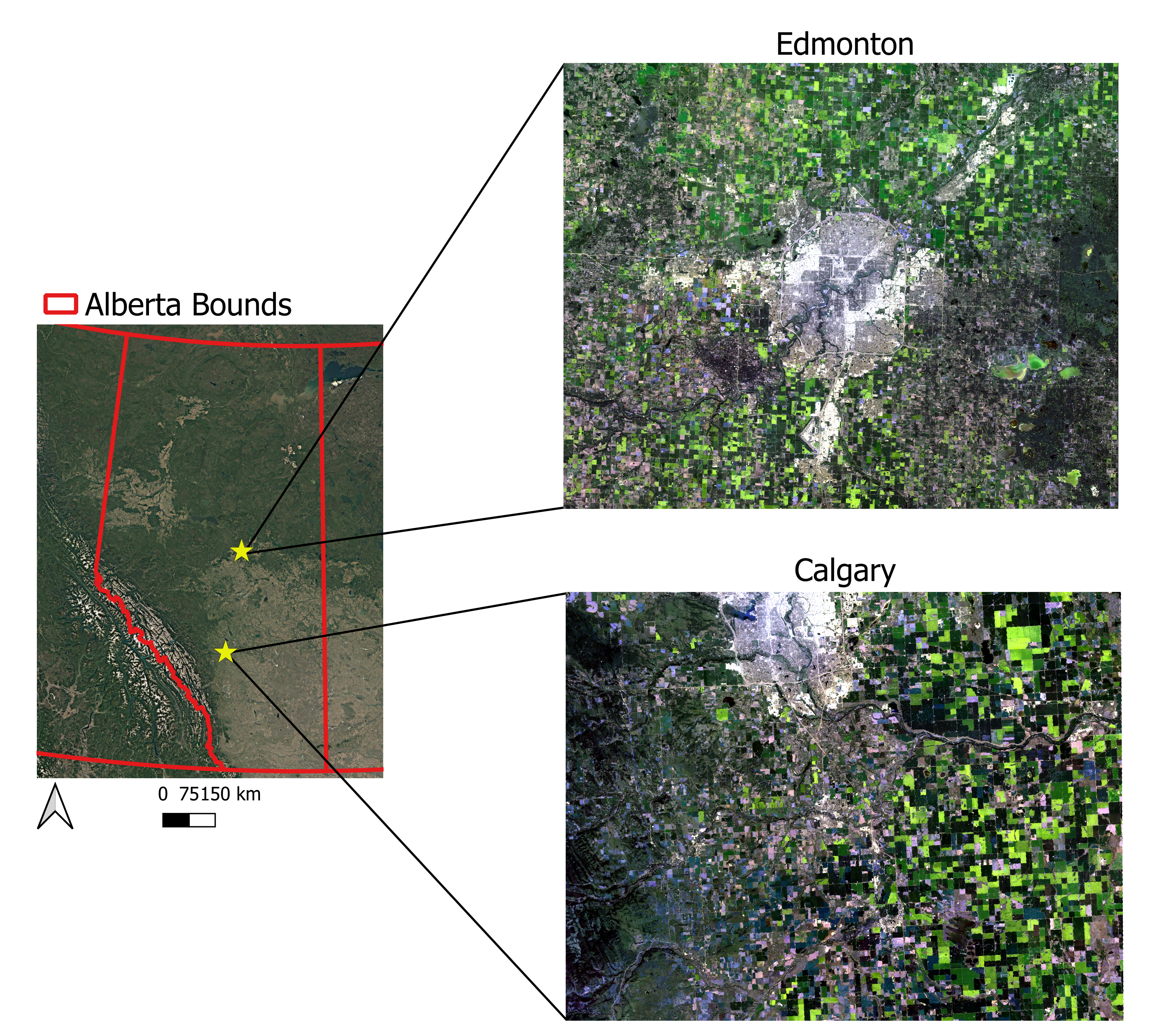}
\caption{Landsat 8 OLI Level-2 imagery of Edmonton (acquired July 21, 2022) and Calgary (acquired July 24, 2022), displayed in UTM Zone 12N as a true colour composition \cite{USGS_Landsat8_2025}.}
\label{fig:calgary-landsat}
\end{figure}

\subsection{Data Collection}

\begin{figure}[ht]
\centering
\includegraphics[width=\linewidth]{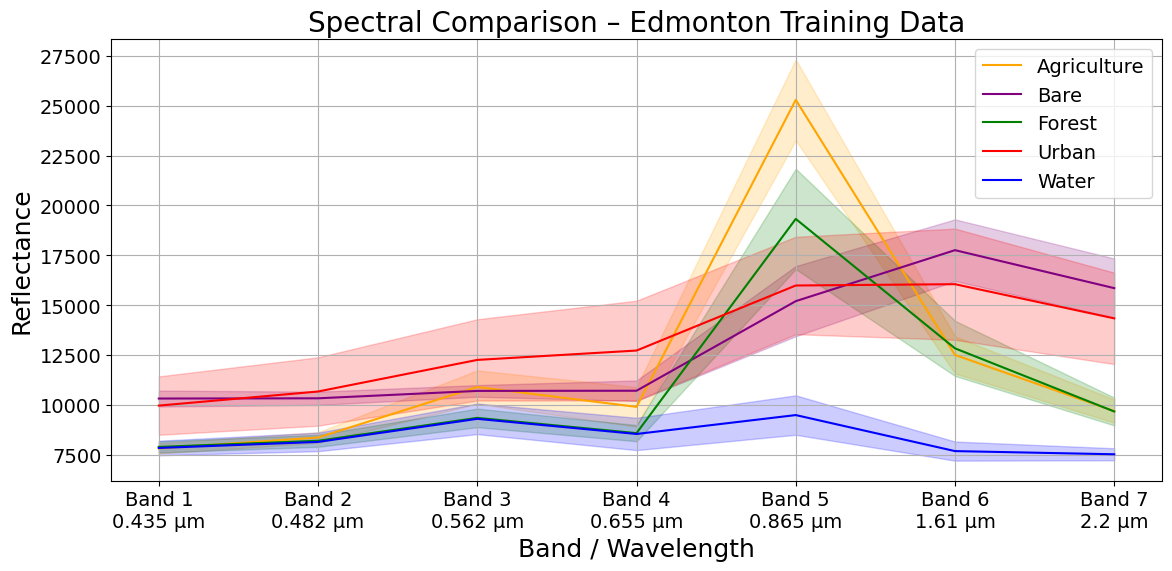}
\caption{Average reflectance and standard deviation of each Landsat 8 band for each respective class.}
\label{fig:spectral-plot}
\end{figure}

Two Landsat 8 OLI Level-2 images from July 2022 \cite{USGS_Landsat8_2025} were used for data collection across five selected LULC classes: agriculture, urban, water, forest, and bare land. A total of 3,000 reference points were manually selected and sampled over the Edmonton area in QGIS (v.3.44.2). The dataset was then divided into into an 80\% training set (2,400 points) and a 20\% held-out validation set (600 points) using stratified random sampling to preserve class balance. Spatial clustering via K-means was applied to ensure that training and validation points were drawn from spatially distinct regions, reducing the risk of spatial autocorrelation inflating validation performance. The mean and median distance between any training and test point was 2,206 m and 2,133 m, respectively. Only 2.8\% of test points were within a 90 m buffer of any training point.
The average reflectance and standard deviation of each data point in the Edmonton training dataset was plotted to visually inspect spectral differences between classes (Figure~\ref{fig:spectral-plot}).

\begin{figure}[ht]
\centering
\includegraphics[width=\linewidth] {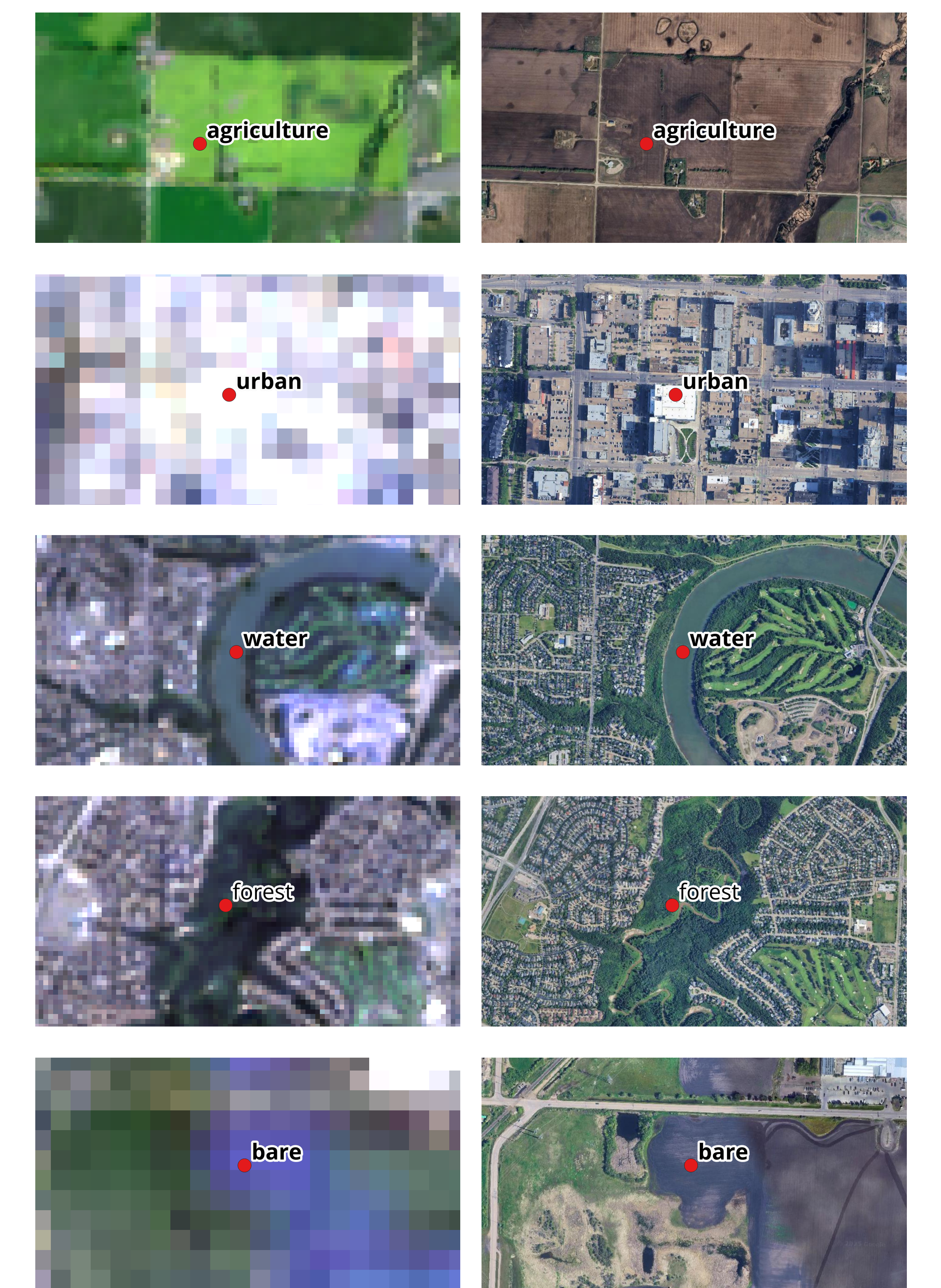}
\caption{Examples of Landsat 8 OLI (left) data points with the Google Satellite overlay (right) for each LULC class.}
\label{fig:data-samples}
\end{figure}

The five LULC classes were selected because they are well represented in both cities. A high-resolution Google Satellite overlay (via QGIS QuickMapServices, EPSG:32612) was used to verify that each point corresponded to a single land class. This visual interpretation approach, while standard in land-cover validation, introduces uncertainty from interpreter subjectivity, temporal mismatch with Landsat imagery, and potential label noise from ambiguous pixels. These factors can affect accuracy assessments. To mitigate this, only clearly homogeneous areas were selected and transitional zones avoided. However, interpreter agreement and label uncertainty were not formally quantified, so reported accuracies should be interpreted with this limitation. Examples of points chosen with Landsat and Google Satellite overlays are shown in Figure~\ref{fig:data-samples}.

A separate set of 600 points was manually collected from Calgary using the same stratified random sampling and visual inspection protocol as Edmonton. These points were used exclusively for evaluating cross-city generalizability and were never accessed during training, validation, or hyperparameter selection.

\subsection{Preprocessing}

Principal Component Analysis (PCA) was performed in \textit{Whitebox Geospatial} (v0.1.1) on Landsat 8 OLI bands 1–7. The first principal component explained 55.5\% of the variance, with the first two and three components accounting for 94.4\% and 99.1\%, respectively. First principal component loadings showed strongest contributions from Bands 5, 6, and 7 (0.705, 0.519, 0.360), followed by Bands 3 and 4 (0.174, 0.213), while Bands 1 and 2 had minimal influence (0.112, 0.128) and were excluded. The remaining five bands were min–max normalized to [0, 1] and used to train, validate, and test the RF, MLP, and KAN models, hereafter referred to as \textit{normalized models}.

For a more comprehensive analysis of how feature selection influences model performance, an additional set of experiments was conducted using combinations of the Soil-Adjusted Vegetation Index (SAVI), Normalized Difference Built-Up Index (NDBI), Normalized Difference Water Index (NDWI), and Band 7 as input features. SAVI enhances vegetation signals while reducing soil brightness effects, particularly in sparsely vegetated areas. NDBI emphasizes built-up regions by contrasting shortwave infrared and near-infrared reflectance. NDWI improves the detection of open water by leveraging strong near-infrared absorption in water bodies. Band 7 (shortwave infrared) provides additional sensitivity to moisture content in soil and vegetation, helping in the discrimination of bare and urban surfaces. Each feature was min-max normalized to a [0, 1] scale to ensure consistency. These models, hereafter referred to as \textit{indices models}, were designed to capture subtle distinctions between vegetated and non-vegetated surfaces, as well as to better discriminate between urban, bare, and water-dominated regions. The application of these indices follow established remote sensing literature \cite{huete1988soil, zha2003ndbi, mcfeeters1996ndwi}.

\subsection{Model Initialization, Optimization, \& Training}
\label{sec:model-init}

All model hyperparameters were tuned exclusively using the Edmonton training and validation subsets, ensuring that no information from the test data influenced model selection. The RF classifier was configured based on four main hyperparameters: the number of decision trees, maximum tree depth, minimum samples per split, and minimum samples per leaf. Grid search with three-fold cross-validation was used to optimize these hyperparameters for both the normalized and indices models, respectively.

The MLP consisted of fully connected hidden layers with ReLU activation functions. Hyperparameters included the number of hidden layers, the number of nodes per layer, learning rate, and the optimizer. Grid search with three-fold cross-validation was used to select the best configuration for normalized and indices models.

This study focuses on a subset of KAN hyperparameters to control functional flexibility and regularization. The optimal KAN models were found using grid search and three-fold cross-validation for the normalized and indices models, targeting the following total loss function

\begin{equation}
\ell_\text{total} = \ell_\text{pred}
+ \lambda \Bigg(
   \lambda_\text{L1} \sum_{l=0}^{L-1} |\Phi_l|_1
   + \lambda_\text{ent} \sum_{l=0}^{L-1} S(\Phi_l)
\Bigg),
\label{eq:loss}
\end{equation}

where $\ell_\text{pred}$ denotes the cross-entropy prediction loss, $\lambda$ controls the overall regularization strength, and $\lambda_\text{L1}$ and $\lambda_\text{ent}$ control the relative contributions of the L1 and entropy regularization terms, respectively. In KAN, $\Phi_l$ denotes the collection of learnable spline-based activation functions in layer $l$. The L1 regularization term is computed by summing the L1 norms of these individual activation functions within a layer, and it penalizes large functional contributions, encouraging sparsity in the set of active edges. The entropy term is computed from the relative L1 contributions of the individual activation functions within a layer and measures how evenly or unevenly functional importance is distributed across edges. Lower entropy corresponds to a situation where only a small subset of activation functions dominates the representation, improving interpretability and reducing redundancy. Together, these regularization terms reduce model complexity and help mitigate overfitting to the training data. For further details, refer to the original paper \cite{liu2024kan}. 

Training and validation of each model was conducted using the Edmonton dataset. The MLP and KAN were trained for a maximum of 300 epochs, with early stopping used to stop training when there was no improvement in performance. Pairwise spatial agreement between each classification output was then quantified by computing the proportion of pixels assigned to the same class for all unique model pairs.

\subsection{Stratified Test Sets}

Following training and model optimization, a separate stratified test set of Edmonton was constructed for each model independently, using the classified raster output of that model as the sampling strata. The number of ground-truth reference points required to achieve a target accuracy of 95\% was estimated using Cochran's formula

\begin{equation}
n = \frac{Z^2 \cdot p_i (1-p_i)}{e^2},
\label{eq:cochran}
\end{equation}

where $n$ is the required sample size, $Z$ is the Z-score for the desired confidence level, $p$ is the maximum variability, and $e$ is the precision. Using a Z-score of 1.96, a maximum variability of 0.5, and a precision of 0.05, the required number of reference points to achieve 95\% accuracy is 385. Points were then allocated proportionally across strata based on the mapped class areas, consistent with design-based accuracy assessment recommendations \cite{olofsson2014good}. This procedure was applied separately for each of the six model configurations, specifically the normalized and indices variants of RF, MLP, and KAN, producing independent, class-proportionally weighted test sets for each.

\subsection{Performance \& Parameter Analysis}

Accuracy assessment followed the framework recommended by the authors in \cite{olofsson2014good}, which provides design-based estimates of overall accuracy, user's accuracy, and producer's accuracy from stratified random samples. Approximate 95\% confidence intervals for overall accuracy were computed using the standard error of a proportion under the stratified design: 

\begin{equation}
    \text{CI}_{95} = \hat{p} \pm 1.96 \sqrt{\frac{\hat{p}(1-\hat{p})}{n}},
    \label{eq:ci}
\end{equation}

where $\hat{p}$ is the estimated overall accuracy and $n$ is the test set size ($n=385$ for Edmonton and $n=600$ for Calgary). Macro-averaged F1 score and Intersection over Union (IoU) are also reported.

The number of trainable parameters was calculated for each optimized model used in the classification process to evaluate model efficiency in relation to accuracy. A separate parameter analysis using the Edmonton indices dataset was conducted to evaluate the performance of each model as the number of trainable parameters increases. To do this, a grid search was performed with the KAN to identify the optimal combination of hyperparameters for each target number of trainable parameters. The total number of trainable parameters of the best KAN model was then computed. Using this as a reference, RF and MLP models were trained via grid search with the constraint that their parameter count had to fall within 100 parameters of the KAN model. Accuracy and the losses for both training and validation were plotted to assess both predictive performance and architectural efficiency across models.

\subsection{Model Interpretability}

Feature attribution for the RF and MLP was computed using Shapley additive explanations (SHAP), which is a model-agnostic framework that quantitatively interprets how input features influence individual predictions \cite{lundberg2017unified}. As a qualitative assessment, a subset of 10 RF trees was plotted, which reduces visual clutter while still capturing structural diversity within the ensemble. For the MLP, a standard feed-forward network was rendered, displaying neurons and connections between layers.

Unlike the RF and MLP, the KAN does not rely on SHAP values for feature attribution. Instead, the KAN has its own internal mechanism for estimating feature importance by quantifying the contribution of each input feature through the learned spline-based activation functions and the associated functional coefficients. These internal importance scores reflect the magnitude of each feature's influence on the learned functional decomposition, which are visualized using the \textit{PyKAN} package.

\section{Results}

\subsection{Model Calibration}

For the RF, the normalized model used 300 trees with a maximum depth of 10, 4 samples per leaf, and 20 samples per split. In contrast, the indices-based model favoured a larger ensemble with shallower trees, using 500 trees with a maximum depth of 5, a minimum of 5 samples per leaf, and 9 samples per split.

The normalized MLP performed effectively with a smaller network of [32, 32], while the indices-based model required a deeper architecture with hidden layer sizes of [128, 128, 64]. Both models used the ReLU activation function, the Adam optimizer, and a learning rate of $1\times10^{-2}$.

The KAN showed an inverse trend compared to the RF and MLP. The normalized model required 10 hidden nodes, whereas the indices model used a smaller architecture of 5 hidden nodes. The learning rate differed slightly, with $1\times10^{-4}$ for the normalized model and $5\times10^{-4}$ for the indices model. All other parameters were consistent across both configurations, including grid size $g=5$, spline order $k=3$, regularization parameters $\lambda=5\times10^{-4}$, $\lambda_\text{L1}=5\times10^{-4}$, entropy coefficient $\lambda_\text{ent}=10$, and LBFGS as the optimizer. 

\begin{figure}[ht]
\centering
\includegraphics[width=\linewidth]{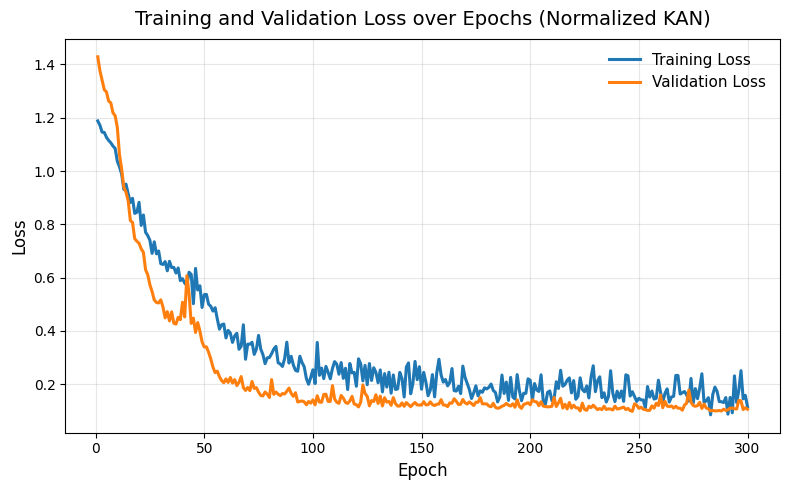}
\vspace{0.2cm}
\includegraphics[width=\linewidth]{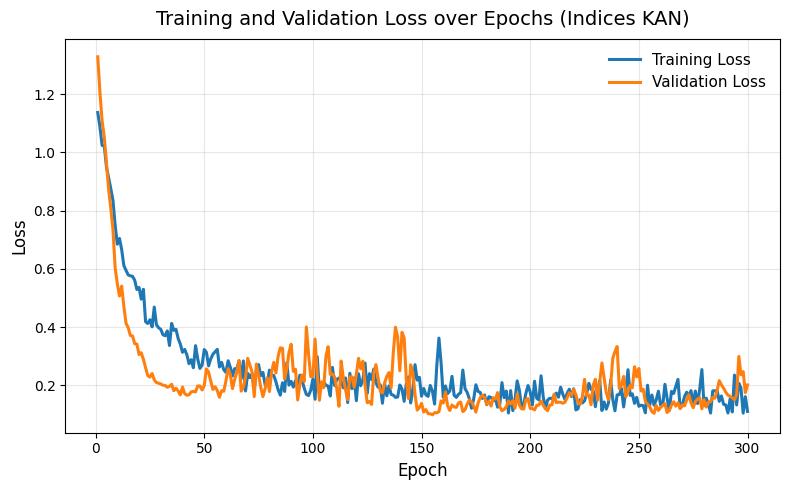}
\caption{Loss curves for the normalized KAN (top) and indices KAN (bottom) models show a smooth decline with minimal separation between training and validation loss, indicating no evidence of overfitting.}
\label{fig:kan-losses}
\end{figure}

During training, the MLP required 100 epochs for the normalized model and 70 epochs for the indices model, while the KAN required 150 epochs for the normalized model and 285 epochs for the indices model. The convergence of the losses in Figure~\ref{fig:kan-losses} confirms that neither KAN model is overfitting to the training data.

\subsection{Classification Outputs}
\label{sec:metrics}

\begin{figure}[!t]
\centering
\includegraphics[width=0.45\linewidth]{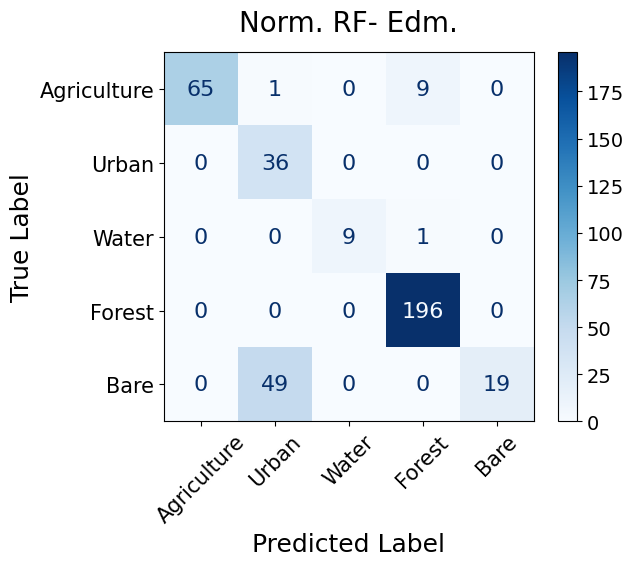}
\label{fig:cm_rf-norm}
\hfil
\includegraphics[width=0.45\linewidth]{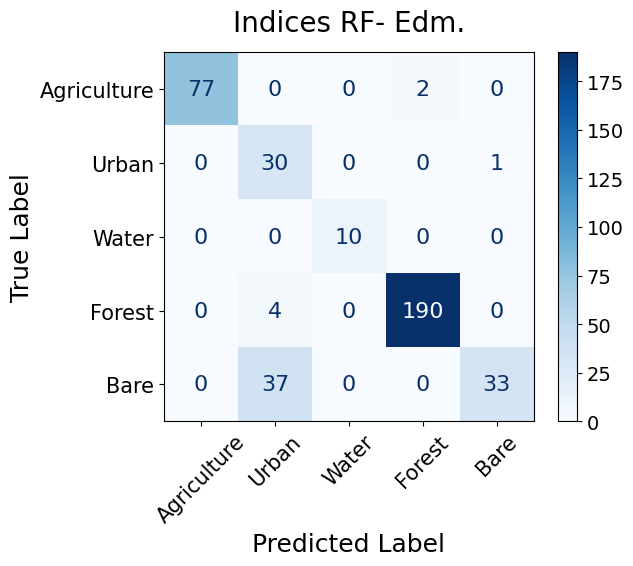}
\label{fig:cm_rf-indices}
\vspace{2mm}
\includegraphics[width=0.45\linewidth]{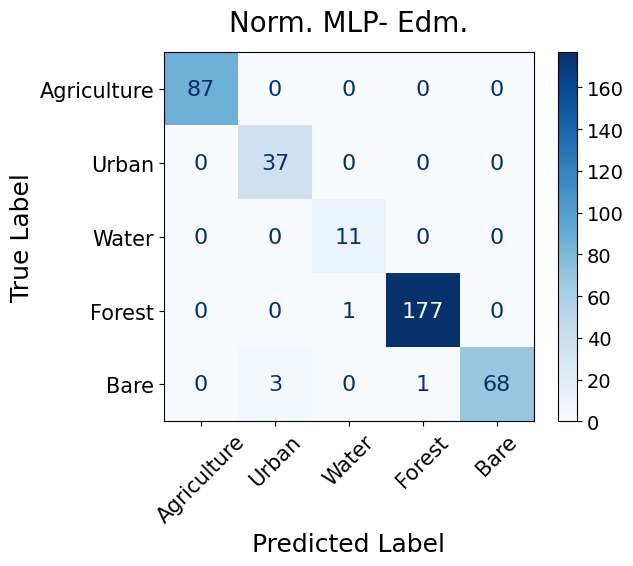}
\label{fig:cm_mlp-norm}
\hfil
\includegraphics[width=0.45\linewidth]{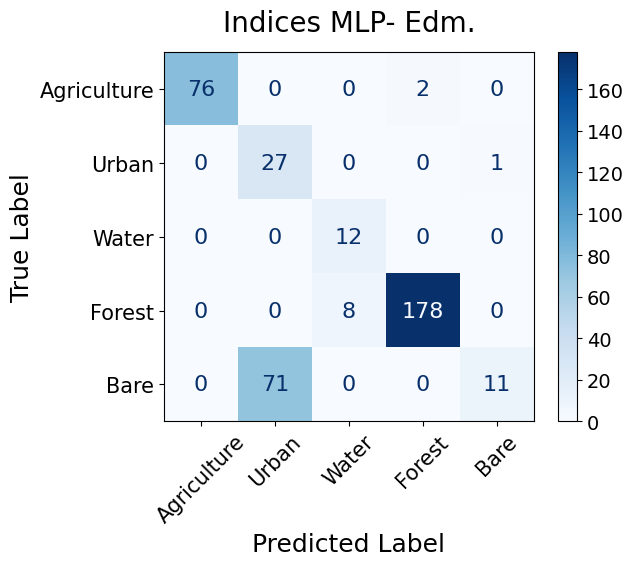}
\label{fig:cm_mlp-indices}
\vspace{2mm}
\includegraphics[width=0.45\linewidth]{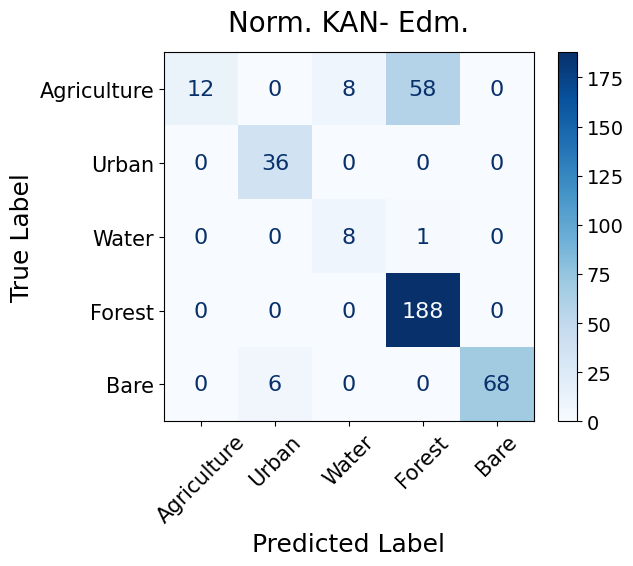}
\label{fig:cm_kan-norm}
\hfil
\includegraphics[width=0.45\linewidth]{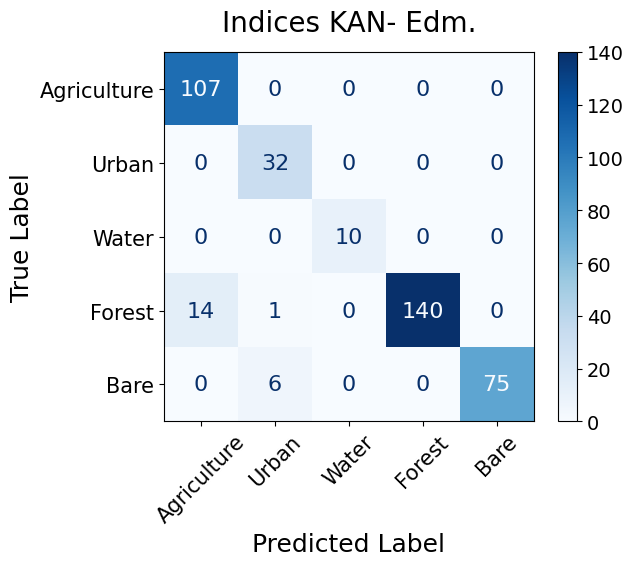}
\label{fig:cm_kan-indices}
\caption{Confusion matrices for each model on the stratified Edmonton test set, where diagonal elements represent correct predictions and off-diagonal elements indicate misclassifications.}
\label{fig:confusion_matrices}
\end{figure}

The error matrices computed on the stratified Edmonton test sets (Figure~\ref{fig:confusion_matrices}) illustrate the class-wise accuracies obtained by each model. Table~\ref{tab:model_comparison} shows the results of the corresponding performance metrics for both cities while Table~\ref{tab:iou_all} shows the IoU values across the five LULC classes for each model.

\begin{table}[!t]
\caption{Classification performance across models, data types, and cities. The highest values of each data type and city are bolded. 95\% confidence intervals are computed using the stratified proportion estimator with n=385 (Edmonton) and n=600 (Calgary).}
\label{tab:model_comparison}
\centering
\footnotesize
\begin{tabular}{llcccc}
\hline
\textbf{Model} & \textbf{City} & \textbf{OA (\%)} & \textbf{$\text{CI}_{95}$ (\%)} & \textbf{F1} \\
\hline
\multicolumn{5}{l}{\textbf{Normalized Bands}} \\
\hline
RF  & Edmonton & 84.4 & [80.8,\ 88.0] & 0.78 \\
MLP & Edmonton & \textbf{98.7} & [97.6,\ 99.8]  & \textbf{0.98} \\
KAN & Edmonton & 81.0 & [77.1,\ 84.9] & 0.73 \\
RF  & Calgary  & \textbf{97.3} & [96.0,\ 98.6]  & \textbf{0.97} \\
MLP & Calgary  & 94.7 & [92.9,\ 96.5] & 0.95 \\
KAN & Calgary  & \textbf{97.3} & [96.0,\ 98.6] & \textbf{0.97} \\
\hline
\multicolumn{5}{l}{\textbf{Indices}} \\
\hline
RF  & Edmonton & 88.5 & [85.3,\ 91.7] &  0.84 \\
MLP & Edmonton & 78.8 & [74.7,\ 82.9] & 0.67 \\
KAN & Edmonton & \textbf{94.6} & [92.3,\ 96.9]  & \textbf{0.95} \\
RF  & Calgary  & 84.3 & [81.4,\ 87.2] & 0.84 \\
MLP & Calgary  & \textbf{89.7} & [87.3,\ 92.1] & \textbf{0.89} \\
KAN & Calgary  & 83.7 & [80.7,\ 86.7] & 0.83 \\
\hline
\end{tabular}
\end{table}

\begin{table}[!t]
\caption{Per-class intersection over union across models, data types, and cities. Bolded values show the highest IoU per data type and city.}
\label{tab:iou_all}
\centering
\footnotesize
\begin{tabular}{llccccc}
\hline
\textbf{Model} & \textbf{City} & \textbf{Agriculture} & \textbf{Urban} & \textbf{Water} & \textbf{Forest} & \textbf{Bare} \\
\hline
\multicolumn{7}{l}{\textbf{Normalized Bands}} \\
\hline
RF  & Edmonton & 0.87          & 0.42          & 0.90          & 0.95          & 0.28          \\
MLP & Edmonton & \textbf{1.00} & \textbf{0.93} & \textbf{0.92}          & \textbf{0.99} & \textbf{0.94}          \\
KAN & Edmonton & 0.15          & 0.86          & 0.47          & 0.76          & 0.92          \\
RF  & Calgary  & 0.97          & \textbf{0.98} & \textbf{0.93} & 0.89          & \textbf{0.98} \\
MLP & Calgary  & \textbf{0.99} & \textbf{0.98} & 0.81          & 0.76          & \textbf{0.98} \\
KAN & Calgary  & 0.98          & 0.97          & 0.91          & \textbf{0.90} & \textbf{0.98} \\
\hline
\multicolumn{7}{l}{\textbf{Indices}} \\
\hline
RF  & Edmonton & \textbf{0.97} & 0.42          & \textbf{1.00} & \textbf{0.97} & 0.46          \\
MLP & Edmonton & \textbf{0.97}          & 0.27          & 0.60          & 0.95          & 0.13          \\
KAN & Edmonton & 0.88          & \textbf{0.82} & \textbf{1.00} & 0.90          & \textbf{0.93}          \\
RF  & Calgary  & 0.90          & 0.63          & 0.91          & 0.61          & 0.61          \\
MLP & Calgary  & \textbf{0.94} & \textbf{0.81}          & 0.75          & 0.62          & \textbf{0.96} \\
KAN & Calgary  & 0.90          & 0.54          & \textbf{0.93}          & \textbf{0.63}          & 0.65         \\
\hline
\end{tabular}
\end{table}

\begin{figure}[!t]
\centering
\includegraphics[width=\linewidth]{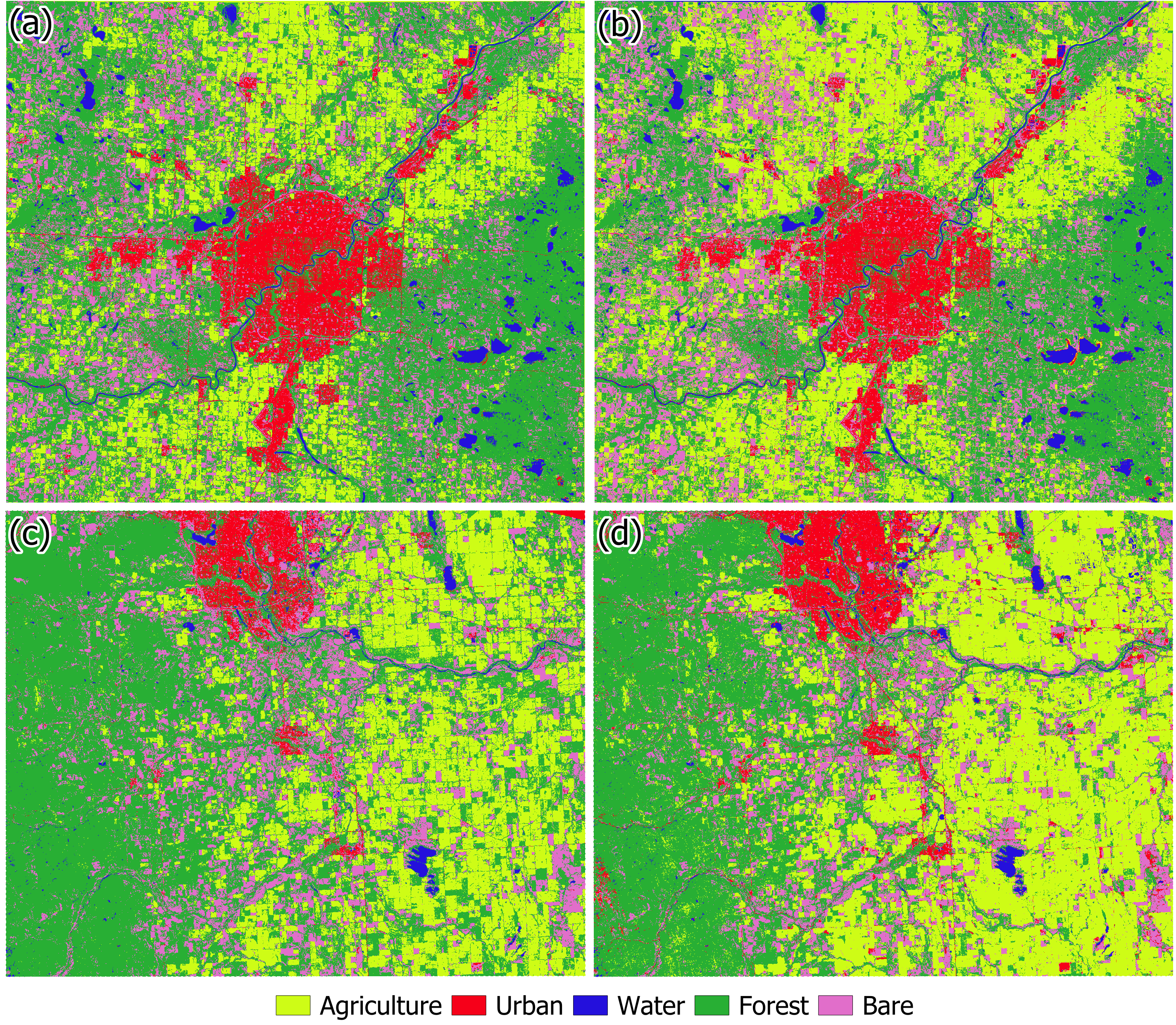}
\caption{Classified Edmonton and Calgary using the different KAN models: Normalized KAN (Edmonton) (a), Indices KAN (Edmonton) (b), Normalized KAN (Calgary) (c), Indices KAN (Calgary) (d).}
\label{fig:edm-class-compare}
\end{figure}

Figure~\ref{fig:edm-class-compare} shows the classification outputs of the KAN models used in this study for both the Edmonton and Calgary regions. The pairwise agreement among all classifications in Edmonton ranged from 84.0\% to 93.2\%, with a mean agreement of 89.1\%, indicating spatial consistency across each model. In Calgary, the pairwise agreement was lower, ranging from 70.3\%-91.2\% with a mean agreement of 82.3\%, reflecting greater variability in model agreement compared to Edmonton.

\subsection{Model Generalization}
\label{sec: generalization}

\begin{figure}[ht]
\centering
\includegraphics[width=\linewidth]{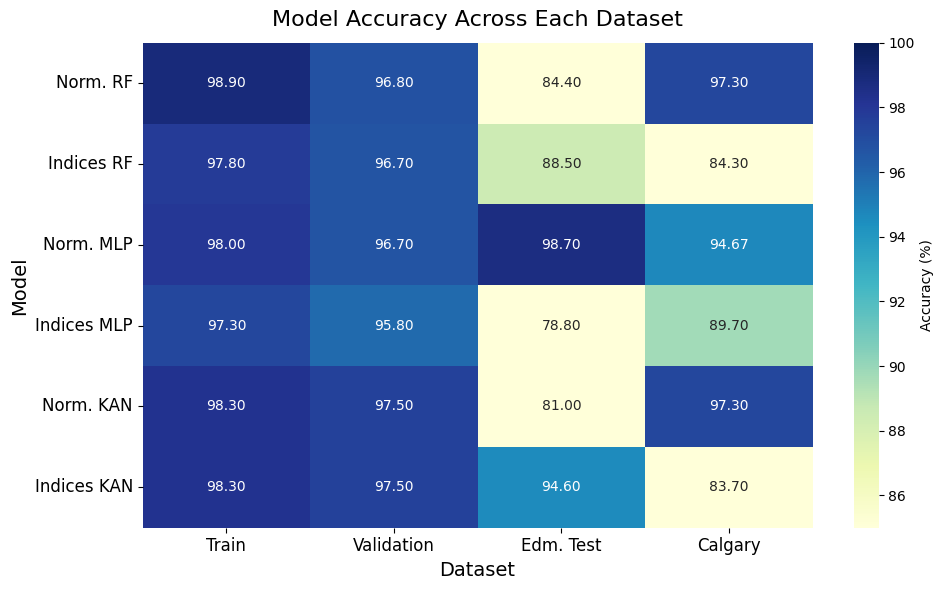}
\caption{Heat map illustrating the generalizability of each model across Edmonton and Calgary datasets.}
\label{fig:heat-map}
\end{figure}

The accuracies of each model for each feature set and city is shown in Figure~\ref{fig:heat-map}. The normalized MLP achieved the best performance on the stratified Edmonton test set with 98.7\% accuracy, while the indices MLP performed the worst with an accuracy of 78.8\%. When generalizing the models to Calgary, both the normalized RF and normalized KAN performed the best with an accuracy of 97.3\%. Meanwhile, the indices KAN performed the worst with an accuracy of 83.7\%.  

\subsection{Parameter Comparison}

The number of trainable parameters was calculated for each optimized model, revealing substantial differences in model complexity. For the normalized inputs, the KAN model used 1,400 parameters, which was significantly lower than the RF with 17,444 parameters and comparable to the MLP with 1,413 parameters. For the indices-based inputs, the KAN model required only 675 parameters, compared to 18,198 for the RF and 25,733 for the MLP.

\begin{figure}[ht]
\centering
\includegraphics[width=\linewidth]{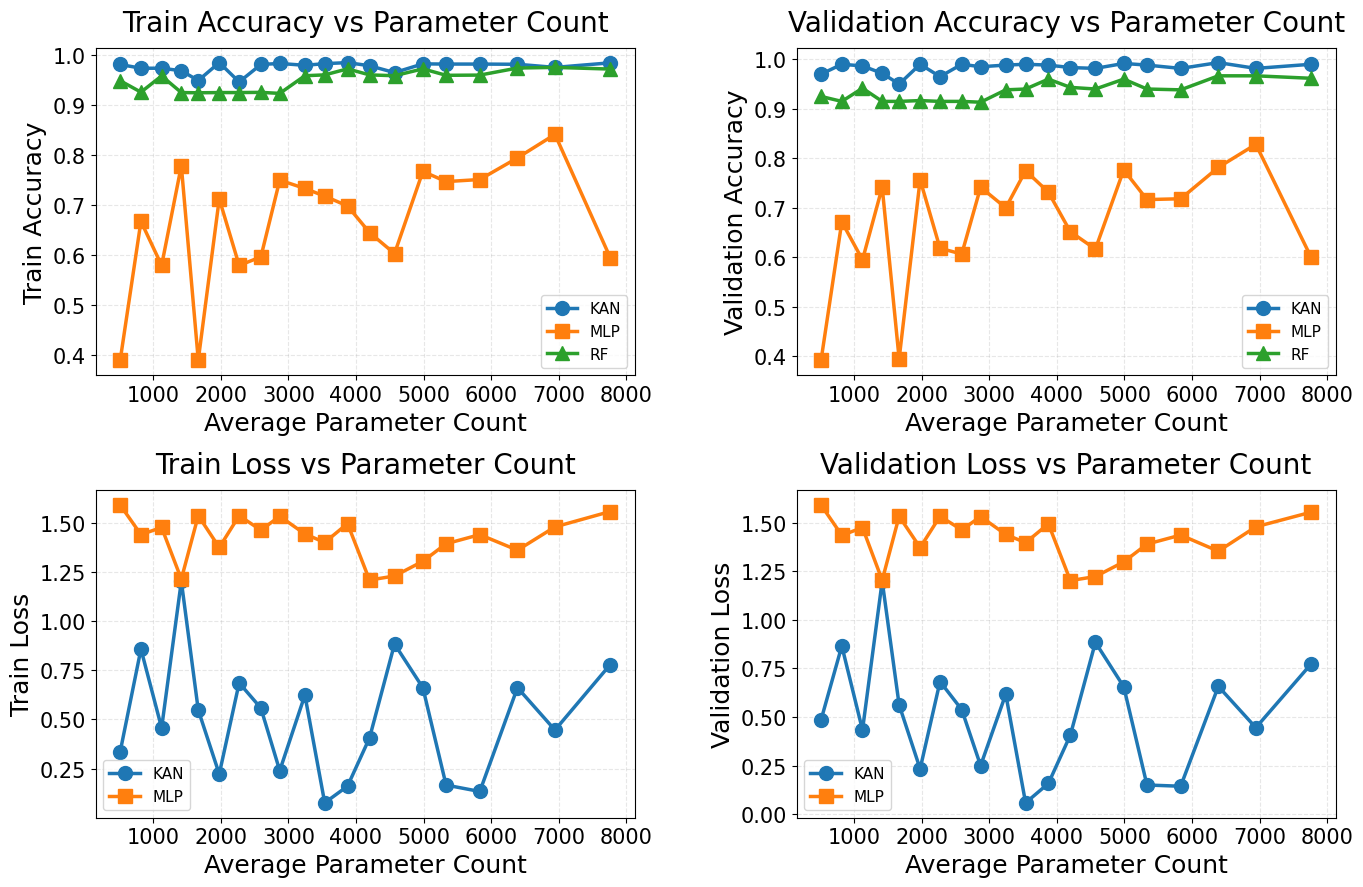}
\caption{A parameter analysis to evaluate each model's architectural efficiency during training and validation on the Edmonton dataset using the spectral indices feature set.}
\label{fig:param-analysis}
\end{figure}

Figure~\ref{fig:param-analysis} shows a comprehensive test of how model performance is affected as the number of trainable parameters increases, hence giving insight to model efficiency. 

\subsection{Interpretability}

\begin{figure}[t]
\centering
\begin{minipage}{0.48\linewidth}
    \centering
    \includegraphics[width=\linewidth]{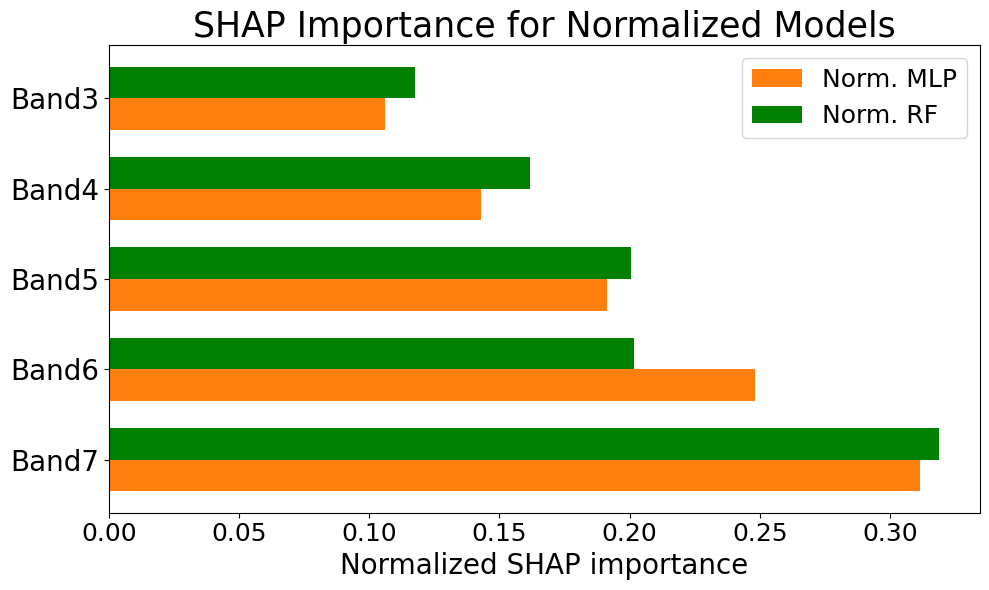}
\end{minipage}
\hfill
\begin{minipage}{0.48\linewidth}
    \centering
    \includegraphics[width=\linewidth]{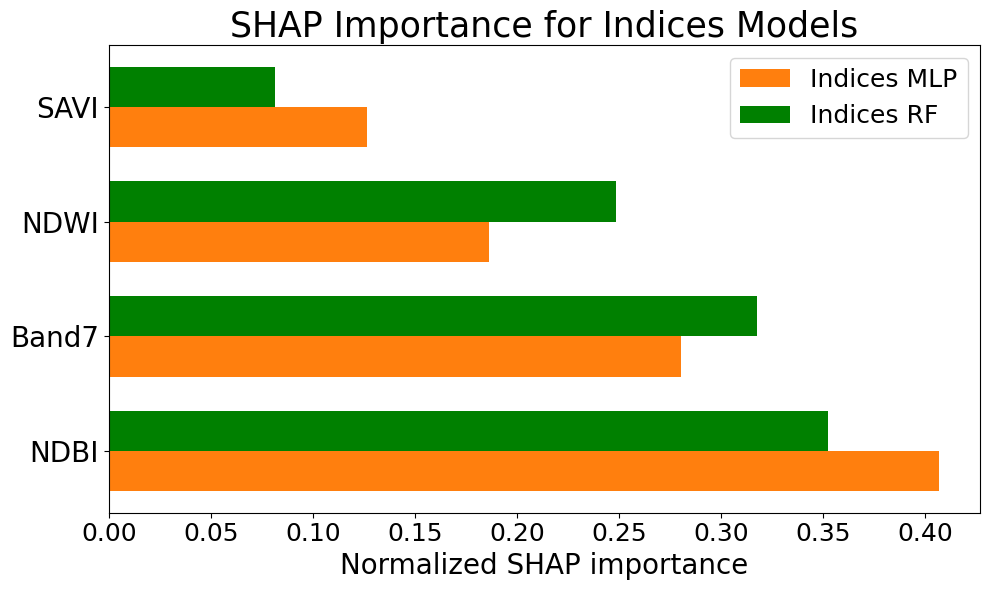}
\end{minipage}
\vspace{0.3cm}
\begin{minipage}{0.48\linewidth}
    \centering
    \includegraphics[width=\linewidth]{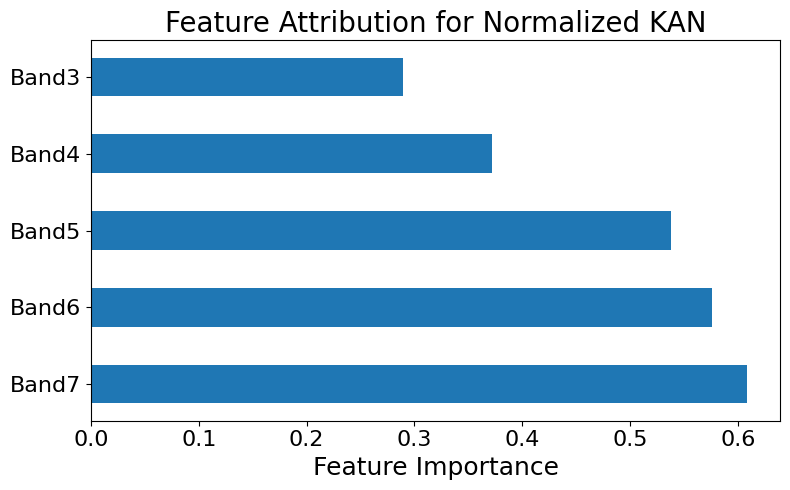}
\end{minipage}
\hfill
\begin{minipage}{0.48\linewidth}
    \centering
    \includegraphics[width=\linewidth]{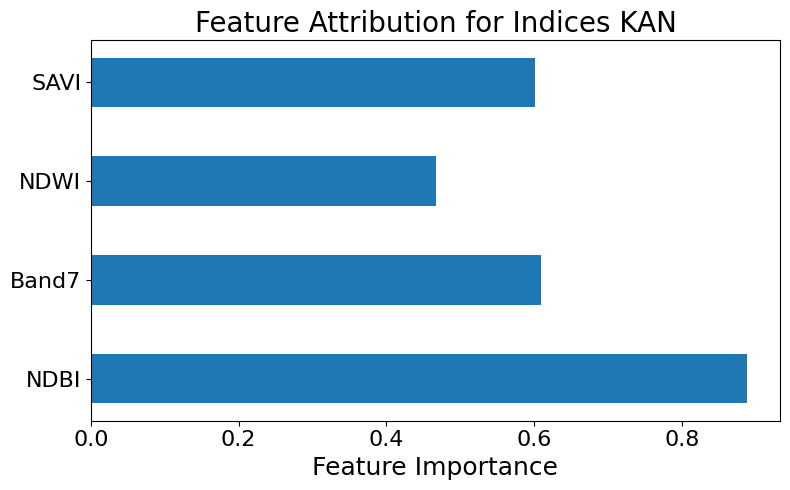}
\end{minipage}
\caption{Feature attribution comparison of each model for the normalized dataset (left column) and the indices dataset (right column).}
\label{fig:feature-interpretability}
\end{figure}

The feature importance of each model is shown in Figure~\ref{fig:feature-interpretability}. Ultimately, the magnitude of the three model's feature attribute values represents the contribution that that specific feature makes towards the predictions. 

\begin{figure}[ht]
    \centering
        \subfloat[Normalized architectures]{%
        \includegraphics[width=0.48\columnwidth]{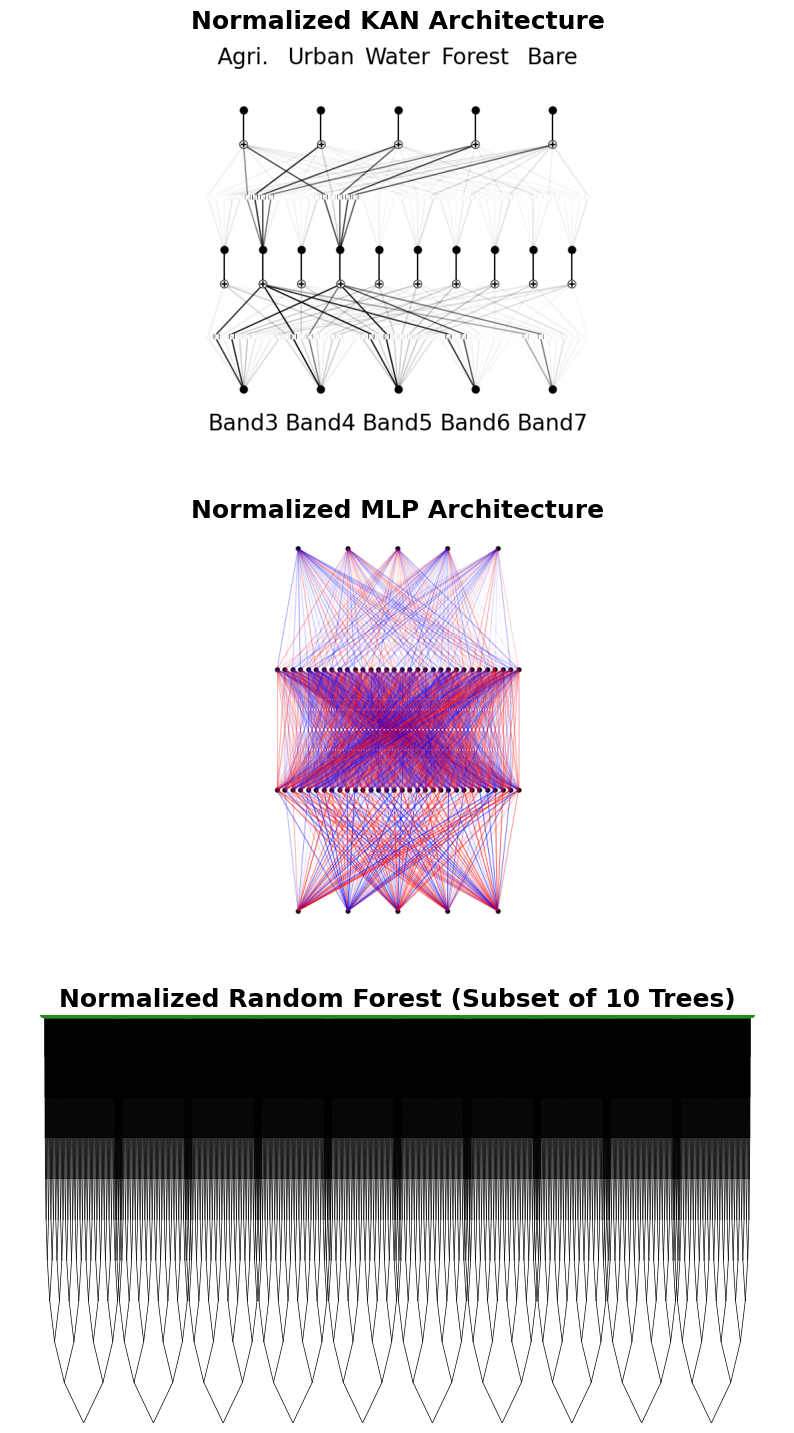}
        \label{fig:norm_arch}
    }
    \hfill
    \subfloat[Indices architectures]{%
        \includegraphics[width=0.48\columnwidth]{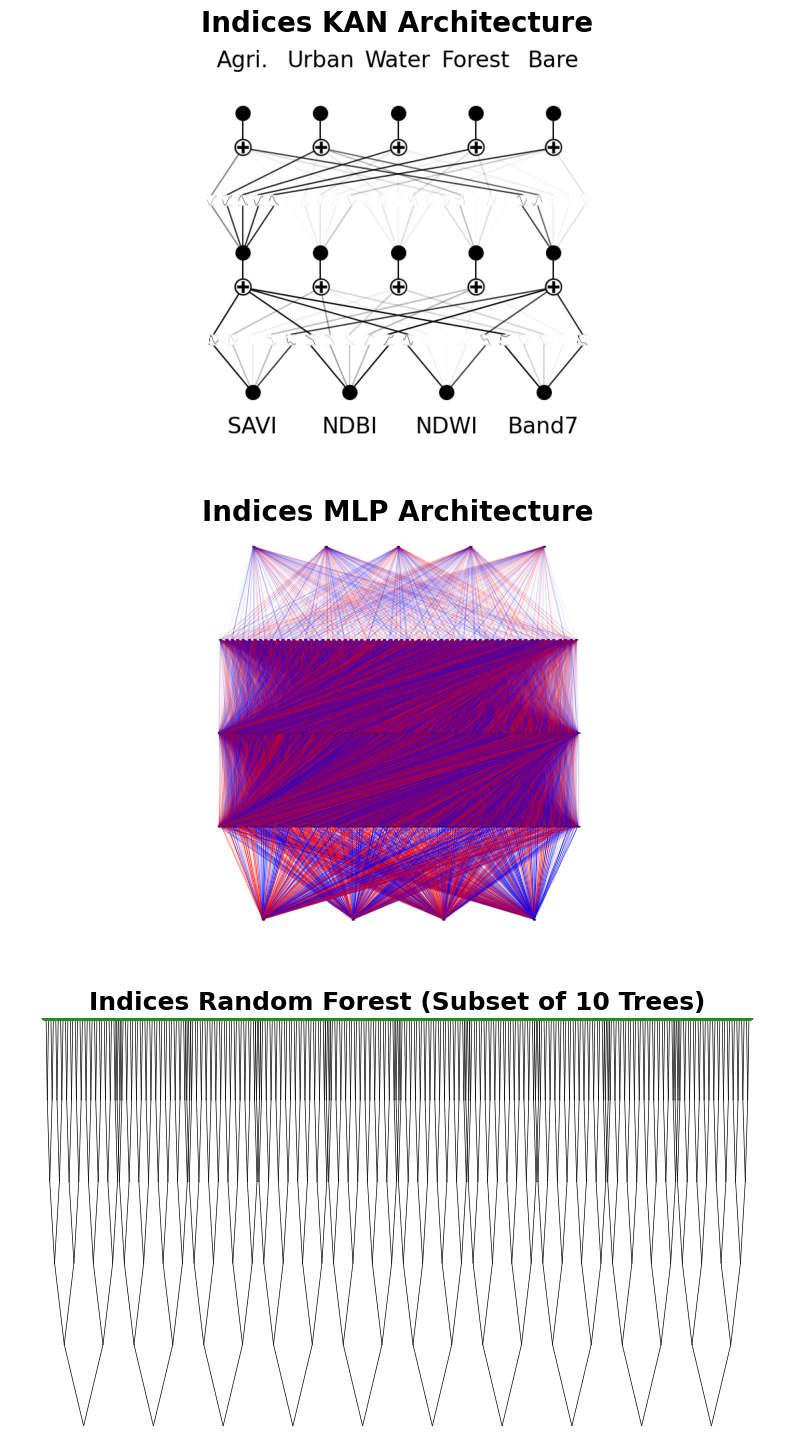}
        \label{fig:indices_arch}
    }
    \caption{Architectures of the RF, MLP, and KAN models used in this study. The KAN visualization highlights learned spline activations and their relative importance as indicated by edge darkness, the MLP displays all node connections, and the RF shows a subset of 10 trees for clarity.}
    \label{fig:combined_architectures}
\end{figure}

The true representation of each model's architecture is shown in Fig.~\ref{fig:norm_arch} and Fig.~\ref{fig:indices_arch}. The RF visualization shows a subsample of 10 trees, the MLP shows each connection between every node in the network, and the KAN shows each connection, learned activation functions, and the importance of each feature.

\section{Discussion}

\subsection{Class-Level Performance and Error Structure}

Both the indices KAN and the normalized MLP achieved perfect class-wise accuracy on the agriculture class, indicating that all agriculture pixels were correctly identified by both models. However, the IoU results in Table~\ref{tab:iou_all} show a difference in spatial agreement, with the normalized MLP achieving a perfect IoU of 1.00 and the indices KAN achieving 0.88. This suggests that although both models correctly classify agriculture samples, the normalized MLP produces a more spatially consistent overlap with the ground truth class boundaries.

The normalized KAN struggled in the same domain. Normalization removes absolute reflectance intensities and narrows the dynamic range across bands 3-7. While normalization benefits generalization, see Section~\ref{sec: generalization}, it reduces the magnitude of spectral differences between soil-heavy agricultural pixels and denser vegetation. This leads to weaker boundary alignment, reflected in the low agricultural IoU. The RF and MLP compensate by using large parameter counts or decision-tree fragmentation while the normalized KAN does not, which may partly explain its relative drop in agriculture performance and spatial consistency in Edmonton. 

Both indices and normalized RF models and the indices-based MLP consistently misclassified bare as urban. This is evident in the Edmonton IoU scores, with values ranging from 0.13 to 0.46, indicating substantial confusion with urban predictions. Bare soil, concrete, and gravel exhibit similar NIR and SWIR reflectance characteristics, which can reduce the discriminative power of indices derived from these bands. As a result, NDBI may assign similarly high values to both bare soil and built-up areas, leading to confusion between these land cover types \cite{oknisia2025effectiveness}. In contrast, both KAN models separate these classes more effectively, achieving much higher bare IoU values of up to 0.93–0.98, depending on configuration. While the mechanism is not explicitly established here, a plausible explanation is that spline-based activations learn smoother nonlinear decision boundaries that preserve subtle spectral differences overlooked by RF and MLP \cite{liu2024kan}. Overall, KAN provides more spatially coherent separation of spectrally similar non-vegetated classes, a known challenge for medium-resolution sensors such as Landsat 8 OLI \cite{oknisia2025effectiveness}.

\subsection{Generalization Across Cities: Radiometry, Scene Heterogeneity, and Model Behaviour}

Both the spectral indices and raw Landsat bands 3–7 were normalized prior to training; however, their cross-scene behavior differs significantly. The reduced performance of the indices-based KAN on Calgary with 83.7\% overall accuracy is consistent with the known sensitivity of spectral indices to radiometric variability between scenes. Indices such as SAVI are not radiometrically invariant, as their values depend on illumination conditions, atmospheric effects, and surface reflectance properties \cite{hadjimitsis2010atmospheric}. When models are transferred across scenes without radiometric correction, these differences can lead to systematic misclassification and degraded land-cover estimates \cite{paolini2006radiometric, song2001classification}. This issue is further amplified by environmental differences between the study areas. Calgary’s higher average elevation of 1045 m relative to 645 m in Edmonton affects atmospheric conditions such as temperature and moisture, shifting baseline reflectance values and causing identical land-cover types to occupy different regions of the index feature space. As a result, the spline functions learned by the indices-based KAN become adapted to Edmonton’s index distributions, effectively overfitting the geometry of that feature space rather than individual samples, which limits transferability. This reflects a broader challenge in multispectral classification, where illumination- and environment-driven spectral shifts violate the assumption of consistent feature distributions across domains and can significantly degrade model performance \cite{chen2025generalizable}. Consequently, while spectral indices can enhance separability within a single scene, they may reduce generalization when cross-scene radiometric variability is substantial.

In our experiments, min-max normalization of the raw bands enabled the KAN to achieve 97.3\% accuracy on Calgary, matching the normalized RF. Min‑max normalization of the raw bands enforces a consistent numerical scale across scenes by rescaling each feature to a fixed range between zero and unity, thereby mitigating absolute brightness and contrast differences that arise from varying illumination and sensor conditions due to factors such as differences in solar zenith angle, atmospheric scattering, and sensor calibration drift. In contrast, spectral indices combine bands nonlinearly, making their distributions highly scene-dependent. Normalizing raw bands directly bounds their absolute values across scenes, preserving relative spectral shapes without reinforcing scene-specific biases. The superior cross-city performance of normalized models is consistent with this reasoning, though a formal normalization ablation or distribution analysis across the two cities was not conducted and would strengthen this interpretation in future work. 

\subsection{Parameter Efficiency and Model Scaling Behaviour}

Comparing trainable parameter counts, the indices KAN required only 675 parameters compared with 18,198 for the indices RF and 25,733 for the indices MLP; the normalized KAN (1,400 parameters) was similarly compact relative to the normalized RF (17,444) and normalized MLP (1,413). However, RF parameter counts are not directly analogous to neural network weights (decision nodes represent discrete binary operations rather than continuous arithmetic) and so this comparison is best interpreted as a structural observation rather than a strict complexity equivalence \cite{breiman2001random}. Runtime efficiency was not benchmarked because the current \textit{PyKAN} implementation is not computationally optimized; training speed is a known limitation acknowledged in the original KAN paper \cite{liu2024kan}. Parameter counts therefore reflect architectural compactness rather than operational efficiency, and runtime benchmarking remains an important direction for future work.

The efficiency of the KAN comes from the structural properties of its functional representation. Rather than relying on large matrices of learned weights, the KAN represents nonlinear transformations using spline basis functions, which can capture complex curvature with a small number of parameters. This reduces the dimensionality of the hypothesis space, allowing the model to express nonlinearities without requiring deep layers or wide networks. Furthermore, KANs learn functions directly rather than combinations of weights, meaning that each parameter has clearer functional meaning and contributes more directly to model expressiveness \cite{liu2024kan}. Additional similar findings have been reported showing that KANs can match or exceed the performance of neural networks while using a small number of parameters, particularly in remote sensing applications \cite{cheon2024kolmogorov}. 

\subsection{Model Interpretability and Structural Transparency}

Although RFs are widely regarded for their strong predictive performance and robustness, their interpretability degrades significantly as the ensemble grows. This decline is primarily due to the heterogeneity and depth of the individual trees within the forest. Each tree is trained on a unique bootstrap sample and a randomly selected subset of features, causing tree structures to diverge substantially. As a result, decision boundaries become highly inconsistent across trees, making it difficult to derive a coherent structural interpretation of the model's behaviour \cite{jamnal2023instils}. 

Even attribution tools such as SHAP provide only a high-level statistical interpretation, indicating the relative contribution of each feature to the overall predictions (seen in Figure~\ref{fig:feature-interpretability}). However, SHAP values cannot uncover the functional mechanisms by which these features interact across the hundreds of branching decision rules in the forest. For example, in this study, SHAP consistently highlighted Band 7 as dominant predictors in both the indices and normalized RFs, but it could not articulate how the feature combines to form decision boundaries between spectrally similar classes such as urban and bare. Consequently, although RFs remain powerful and practical tools for land-cover classification, their interpretability remains inherently limited due to the disordered and opaque structure of their constituent trees. 

MLPs offer complete architectural visibility because every node and edge is directly accessible, as seen in Figure~\ref{fig:combined_architectures}. However, this does not equate to interpretability in any meaningful sense. The transformations occurring within the hidden layers are mathematically well-defined but semantically opaque. In land-cover classification, this opacity presents a major challenge: although the network may achieve high predictive performance, the internal logic remains inaccessible. 

The KAN's graphical representation of edge thickness (as determined by functional magnitude) in Figure~\ref{fig:combined_architectures} conveys how strongly each input contributes to each intermediate representation, offering an intuitive, mechanistic understanding of the learned relationships. Although interpretability declines as model complexity increases, particularly when moving from the indices-based to the normalized KAN, the architecture remains more transparent than both RF and MLP models.

\subsection{Implications for Mapping Urban Regions}

The findings of this study suggest that KANs are a viable option for LULC classification workflows, though their advantages over RF and MLP models are not consistent across all conditions. In Edmonton, the normalized MLP outperformed the normalized KAN by a notable margin with 98.7\% versus 94.6\% overall accuracy, indicating that KANs do not universally match or exceed high-parameter networks on within-scene accuracy. In Calgary, the normalized KAN and normalized RF achieved equivalent overall accuracy at 97.3\%, suggesting competitive generalization under some conditions but not a clear advantage for KANs specifically. Performance was also sensitive to feature representation: index-based KANs underperformed normalized KANs in Calgary, while normalized KANs underperformed the normalized MLP in Edmonton. These results highlight that the suitability of KANs depends on both the geographic context and the chosen feature configuration.

The compact KAN architecture remains a practical advantage in resource-constrained settings. KANs achieved competitive accuracy using substantially fewer trainable parameters than the MLP, making them a reasonable candidate for deployment on satellites, unmanned aerial vehicles, and edge devices where computational overhead is a limiting factor. Overall, KANs represent a competitive and interpretable option for satellite-based land classification, though claims of general superiority in accuracy are not supported by the present results.

Finally, KANs do introduce a meaningful shift in model interpretability for remote sensing. Unlike RFs and MLPs, which offer limited insight into their internal decision-making, KANs provide a direct mapping between input spectral values and the output class via visible functional transformations. This transparency allows inspection of how vegetation indices, SWIR reflectance, and other spectral dimensions influence classification, which may guide field campaigns, ground truth collection, and sensor calibration regardless of whether KANs achieve the highest accuracy in a given setting. 

\subsection{Limitations}

Although the results demonstrate certain advantages to using the KAN for land classification, several limitations should be acknowledged. The analysis is restricted to two cities within the same climatic region of western Canada. While Edmonton and Calgary exhibit differences in elevation, surface materials, and atmospheric conditions, they do not capture the full range of spectral, phenological, and land-use variability encountered globally. Consequentially, the reported generalization performance may not extend to regions with substantially different environmental or urban characteristics.

The study is based on single-date Landsat 8 OLI imagery, which does not account for seasonal variability or phenological dynamics. This may limit robustness for vegetation-related classes where spectral responses change over time. In addition, the feature space is intentionally constrained to a small set of spectral bands or commonly used indices, excluding texture measures, topographic variables, and physically based features that could further improve class separability and cross-scene stability. Finally, model efficiency is evaluated primarily through parameter counts rather than training or inference time, leaving practical deployment considerations unaddressed.

\section{Conclusion}

The results demonstrate that model performance varied depending on both the study region and feature representation. While the MLP achieved the highest overall accuracy of 98.7\% in Edmonton, the KAN matched the RF in Calgary with an overall accuracy of 97.3\%. Although KAN did not consistently outperform competing models, it achieved strong classification performance using substantially fewer trainable parameters and offered improved interpretability through its learnable univariate functions. These findings indicate that KAN is a viable and efficient approach where model compactness and interpretability are important. 

This study highlights several directions for advancing KAN architectures in remote sensing. First, expanding the spatiotemporal scope beyond two Canadian metropolitan regions to include diverse eco regions, surface types, and atmospheric conditions would allow rigorous generalization evaluation of KANs across broader biomes.

Second, architectural refinement through hybrid approaches, such as integrating convolutional priors or transformer-based spatial encoders with KAN functional layers, such as in CNN-KAN and ViT-KAN, could capture both pixel-level spectral nuances and neighborhood-level spatial patterns, enhancing accuracy and generalization without sacrificing interpretability.

Third, incorporating physics-informed features and atmospheric correction parameters, such as surface albedo, leaf area index, moisture indices, and bidirectional reflectance distribution function information, could help disambiguate spectrally similar classes and mitigate cross-scene radiometric variability, further strengthening model generalization.

Finally, domain-adaptation strategies tailored to KANs could adjust spline functions in response to domain shift, enabling dynamic recalibration to new scenes while preserving interpretability and compactness. Collectively, these directions point toward developing KANs as interpretable, physics-informed, and spatially aware models for next-generation remote-sensing applications.

\section*{Acknowledgments}

This research was funded by the Government of Canada's NSERC USRA and the Mitacs Globalink Grant. We would like to thank Patrick O'Brien for his guidance on implementing the KAN and for providing a preliminary paper review; critiques and ideas offered by the Centre for Earth Observation Sciences (CEOS) lab at the University of Alberta; and the collaborations at the Finnish Centre for Artificial Intelligence (FCAI) at Aalto University, Helsinki, Finland. 

\bibliographystyle{IEEEtran}
\bibliography{refs}

\end{document}